\documentclass[a4paper,12pt]{article}

\usepackage{arxiv}

\usepackage[utf8]{inputenc} 
\usepackage[T1]{fontenc}    
\usepackage{hyperref}       
\usepackage{url}            
\usepackage{booktabs}       
\usepackage{amsfonts}       
\usepackage{nicefrac}       
\usepackage{microtype}      
\usepackage{lipsum}

\usepackage[pdftex]{graphicx}

\graphicspath{{noiseimages/}}
\usepackage{enumitem}
\usepackage{amssymb}
\usepackage[nooneline]{caption}

\usepackage{blindtext}
\usepackage{scrextend}
\addtokomafont{labelinglabel}{\sffamily}

\usepackage{amsmath}
\usepackage{enumerate}
\usepackage{multirow}

\title{A Method for Estimating the Proximity of \\ Vector Representation Groups in \\ Multidimensional Space. \\  On the Example of the Paraphrase Task.}

\author{
  Artem~Artemov,\ Boris~Alekseev
   \\
  Cognitive Systems Company\\
  \texttt{science@cogsys.company} \\
}

\begin{document}
\maketitle


\vspace{1cm}

\begin{abstract}
The following paper presents a method of comparing two sets of vectors. The method can be applied in all tasks, where it is necessary to measure the closeness of two objects presented as sets of vectors. It may be applicable when we compare the meanings of two sentences as part of the problem of paraphrasing. This is the problem of measuring semantic similarity of two sentences (group of words). The existing methods are not sensible for the word order or syntactic connections in the considered sentences. The method appears to be advantageous because it neither presents a group of words as one scalar value, nor does it try to show the closeness through an aggregation vector, which is mean for the set of vectors. Instead of that we measure the cosine of the angle as the mean for the first group vectors projections (the context) on one side and each vector of the second group on the other side. The similarity of two sentences defined by these means does not lose any semantic characteristics and takes account of the words’ traits. The method was verified on the comparison of sentence pairs in Russian. 
\end{abstract}

\section*{Introduction}

Text \textit{similarity} task implies determining how similar two pieces of text are in lexical (word level similarity) and semantic (phrase paragraph level similarity) terms. Text similarity measures are being increasingly applied in text related research and tasks such as text classification, text summarization, document clustering, topic detection, topic tracking, questions generation, question answering, short answer scoring, machine translation, and others. Despite the fact that there is a variety of \textit{text similarity measures}, it should be mentioned that some of them appear to be more efficient than others. Down below we provide a short overview of the most popular ways of computing sentence similarity \cite{5}.

\begin{itemize}
\item[$\bullet$] Jaccard Similarity (Jaccard similarity or intersection over union is size of intersection divided by size of union oftwo sets, although it is to be mentioned that the score will be small if there are few common words in the sentences);
\item[$\bullet$] Different embeddings + K-means (With K-mean related algorithms, we first need to convert sentences into vectors, which can be done with Bag of words with either TF (term frequency) called Count Vectorizer method, TF-IDF (term frequency - inverse document frequency) or Word Embeddings.  They  may come from pre-trained methods such as Fastext, Glove,Word2Vec, customized method by using Continuous Bag of Words (CBoW)/ Skip Gram models);
\item[$\bullet$] Different embeddings + Cosine Similarity (Cosine similarity calculates similarity by measuring the cosine of angle between two vectors. The studies show that using Count Vectorized method together with cosine similarity gives less trustworthy results than using pre-trained methods such as Glove with cosine similarity);
\item[$\bullet$] Different embeddings + LSI (Latent Semantic Indexing) + Cosine Similarity (It is used to reduce the dimensionality of our document vectors and keeps the directions in our vector space that contain the most variance);
\item[$\bullet$] Different embeddings + LDA + Jensen-Shannon distance (It is an unsupervised generative model that assigns topic distributions to documents, which also known as information radius (IRad) or total divergence to the average);
\item[$\bullet$] Different embeddings+ Word Mover Distance (It uses the word embeddings of the words in two texts to measure the minimum distance that the words in one text need to “travel” in semantic space to reach the words in the other text);
\item[$\bullet$] Different embeddings + Variational Auto Encoder (VAE) (It encodes data to latent (random) variables, and then decodes the latent variables to reconstruct the data);
\item[$\bullet$] Different embeddings + Universal sentence encoder (Pre-trained sentence encoders have a similar role – as it is with Word2vec and GloVe, but for sentence embeddings: the embeddings they produce can be used in a variety of applications, such as text classification, paraphrase detection, etc.);
\item[$\bullet$] Different embeddings + Siamese Manhattan LSTM (Siamese networks perform well on similarity tasks and have been used for tasks like sentence semantic similarity, recognizing forged signatures and some others);
\item[$\bullet$] Knowledge-based Measures (Knowledge-based measures quantify semantic relatedness of words using a semantic network and shows efficient results)
\end{itemize}

It is widely acknowledged that it is effective to vectorize two words while measuring semantic similarities of those. For instance, it can be done with the model word2vec. However, vectorizing a phrase or a sentence while comparing two sentences might lead to a loss of semantic information. Such a form of vectorizing has a bigger dimensionality rather than that of a word. The semantic „closeness“ of the sentences is defined as „the closeness“ o their vectors. It means that two sentences with similar lexical units but different word order will be treated in a similar way.

In the present paper the way of measuring similarity of two sentences is defining the cosine of the angle between them. The cosine of the angle between a vector and a set of vectors is calculated as the cosine of the angle between the vector and its projection on the plane. The plane is formed by vectors of the set and their linear combinations.

The method is considered to be advantageous because it can be combined with existing methods that use vectors. For instance, it is possible to use not only semantic embeddings but also add syntactic and morphological parts to the process of measuring.

\section{Problem Statement}

There is a matrix
\begin{equation}
M_{ij}=
\begin{bmatrix}
   &\mathbf C_1       & \mathbf C_2   & \vdots & \mathbf C_j \\
     \mathbf F_1 & w_{11}       & w_{12}  & \vdots & w_{1j} \\
     \mathbf F_2 & w_{21}       & w_{22}  & \vdots & w_{2j} \\
    \vdots & \vdots & \vdots & \vdots \\
    \mathbf F_i  & w_{i1}       & w_{i2}  & \vdots & w_{ij}
\end{bmatrix}, \overrightarrow F_i^T  \cdot  M_{ij}  =\overrightarrow C_j
\end{equation}

$F$ - Features, 
i - weight, 
$w \in \mathbb{R}$,  
vector $C$ - Classes j base a basis $ C_j =  \{ e_1 \dots e_i \} ,  e_i \in \mathbb{B}$ and Collections
\begin{equation}
Coll_1= \{ \overrightarrow F_k \}
\end{equation}
\begin{equation}
Coll_2= \{ \overrightarrow F_m \}
\end{equation}

Required to offer a measure of similarity $Coll_1$ and $Coll_2$ that have the following properties 

\begin{itemize}
\item[-] invariance
\item[-] geometric interpretation
\item[-] computability
\item[-] verifiability (comparability with the result of human evaluation)
\end{itemize}

 $SIM (Coll_1; Coll_2) = ?$

Validation of the proposed SIM evaluation method ($Coll_1; Coll_2$) is based on the estimates obtained for the paraphrase classification task \footnote{Paraphrase classification competition track. URL: http://paraphraser.ru/blog}.
 Within the framework of the introduced axiomatics, the “cognitive task” is as follows. There are two texts, which are represented as a sequence of words. Words of one sentence are grouped according to a set of criteria. Each word has a vector representation. Vector’s components are real numbers ("coordinates"). When we are required to compare two texts, it means to offer a certain quantitative characteristic (coefficient) of texts proximity. It is desirable that the proximity coefficient takes values from 0 to 1, and increases when choosing more "similar" texts. The case of a full match should be represented by value 1.

\subsection{Vector methods}

One of the methods for estimating the proximity of the vectors $a, b$ is the cosine of the angle between them

\begin{equation}
cos(a,b) = \frac {a \cdot b} {|a||b|}
\end{equation}

To compare two groups of vectors  $a=(a_i, i=1..n), b=(b_j, j=1..m)$  we can calculate the cosine for each pair of vectors $a_i, b_j$ and then use the average of all such pairwise cosines to estimate the proximity of the groups $a, b$.

As such, it is of interest to generalize the concept of cosine, so that it can be used to estimate the proximity of a vector to a certain group of vectors.

\section{Proposed solution}

Let $b$ be some row-vector, $a$ - a matrix, each row of which is a row-vector of a group $a$.

We propose as angle between vector $b$ and the group $a$ to take the angle between vector $b$ and its projection $p$ on the plane formed by the linear combinations $\lambda a$ of vectors $a$ (here $\lambda$ is a row-vector of arbitrary real coefficients of a linear combination).

\begin{figure}[h]
\begin{minipage}[h]{0.47\linewidth}
\center{\includegraphics[width=1\linewidth]{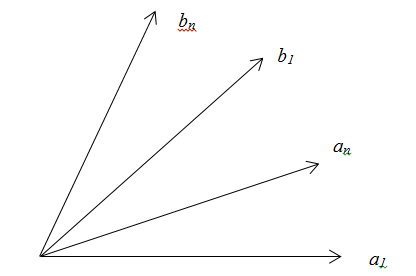}} \\1) Vectors $a_1, \dots, a_n ; b_1 \dots b_m $
\end{minipage} 
\hfill
\begin{minipage}[h]{0.47\linewidth}
\center{\includegraphics[width=1\linewidth]{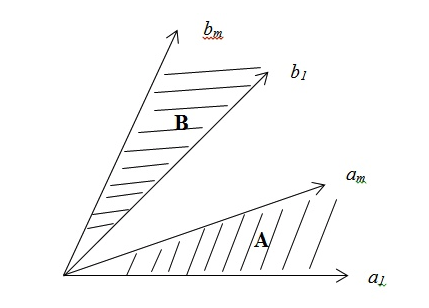}} \\2) Planes of vectors groups  $a, \ b$
\end{minipage}
\vfill
\begin{minipage}[h]{0.42\linewidth}
\center{\includegraphics[width=1\linewidth]{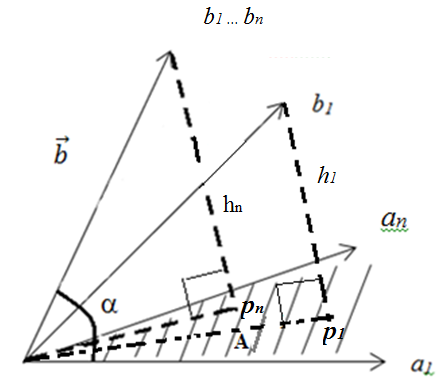}} \\3) Projection $\vec{p}$ of vector $\vec{b}$ on the plane  $A, \vec{h}=\vec{b} - \vec{p}$
\end{minipage}
\hfill
\begin{minipage}[h]{0.45\linewidth}
\center{\includegraphics[width=1\linewidth]{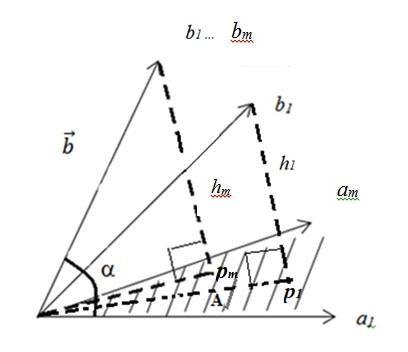}} \\4) Projection $\vec{p}$ of vector $\vec{a}$ on the plane  $B, \vec{h}=\vec{a} - \vec{p}$
\end{minipage}
\label{ris:experimentalcorrelationsignals}
\end{figure}

Cosine of the angle between vector $\vec{b}$ and its projection $\vec{p}$ equals $\frac{|\vec{p}|}{|\vec{b}|}$ - the ratio of the length of the adjacent leg and the hypotenuse $\vec{b}$
\begin{equation}
\vec{h}=\vec{b}-\vec{p}
\end{equation}

\subsection{Alekseev’s method}

Let us derive the cosine angle formula between vector $b$ and its projection $C$ on plane $A$. Vector $h = b-C$ is a normal vector omitted from the end of vector $b$ on plane $A$. It is orthogonal to the vector-rows of the matrix $A$. The projection vector $C$ lies in the plane $A$, which means it is a linear combination $C = \lambda A$ of the vectors of this plane.

According to these two facts, the vector of the $\lambda$ coefficients of this linear combination is determined from the following equations
\begin{equation}
(\vec{b} - \vec{p}) \cdot A^T = (\vec{b} - \lambda A) \cdot A^T = 0
\end{equation}
\begin{equation}
\lambda \cdot A \cdot A^T = \vec{b} \cdot A^T
\end{equation}

If a square matrix
\begin{equation}
A \cdot A^T
\end{equation}

is reversible (that is, when the rows of the matrix $A$  are linearly independent), vector of the coefficients $\lambda$ is determined by the formula:
\begin{equation}
\lambda = \vec{b} \cdot A^T (A \cdot A^T)^{-1},
\end{equation}

and the projection vector $p$ is determined by the formula:
\begin{equation}
\vec{p} = \lambda A = \vec{b} \cdot A^T (A \cdot A^T)^{-1} \cdot A
\end{equation}

We would need the length of the vector $p$. Its square is equal to
\begin{equation}
|\vec{p}|^2 = \vec{p} \cdot \vec{p}^T = \vec{b} \cdot A^T (A \cdot A^T)^{-1} \cdot A \cdot (A^T (A \cdot A^T)^{-1} \cdot A \cdot \vec{b}^T) = \vec{b} \cdot A^T (A \cdot A^T)^{-1} \cdot A \cdot \vec{b}^T
\end{equation}

Thus, the cosine of the angle between vector $b$ and group $A$, defined as the cosine of the angle between vector $b$ and projection $p$ on the plane $A$, 
is equal to the ratio of the projection length $|p|$ to the length of vector $b$
\begin{equation}
cos(\vec{b},A) = cos(\vec{b},\vec{p}) = \dfrac{|\vec{p}|}{|\vec{b}|}
\end{equation}

Substituting the expression for the length $|p|$ of the vector $p$, we finally obtain
\begin{bf}
\begin{equation}
\boldsymbol{cos(\vec{b},A) = \dfrac{ \sqrt{\vec{b} \cdot A^T (A \cdot A^T)^{-1} \cdot A \cdot \vec{b}^T}}{|\vec{b}|}}
\end{equation}
\end{bf}

Using the proposed cosine formula between the vector and the group, we can calculate the cosine between two groups of vectors, understanding it as the average value of the cosine of the first group vector with respect to the entire second group. Note that the so defined measure of closeness $cos (A, B)$ of two groups of vectors $A, B$ is noncommutative and, moreover, generally speaking, depends on the composition of the group vectors. Noncommutativity can be eliminated by symmetrization: $SIM(A, B) = \dfrac{(\sum_i cos (b_i, A) + \sum_j cos (a_j, B))}{2}$. The dependence of cosine on the groups’ composition is limited. Namely, the cosine of groups does not change if the groups do not change their basis of linearly independent vectors (for example, if the groups linearly dependent on the basis of the vector are removed from the group or added to it).

For the two groups consisting of $n$ and $m$ vectors, this will no longer require $\dfrac{nm}{2}$ calculations, but only $m$. However, to ensure the symmetry of the two groups in assessing their proximity, you can complement this by calculating the cosine of each vector of the second group relative to the entire first group, and then take the average of these two proximity measures. This will require $n + m$ cosine calculations.

\subsection{The semantic cores method}

The method is partially described in the paper by Artemov et al \cite{1}. The essence of the method is to compile an on-the-fly neural network linking the word embeddings (semantic content) to classes that reflect their possible expression for the problem solved (for example, sentence selection). In this case, the weights in this network are represented by a sum of weights (for each vector representation with respect to the class under consideration) from the Big Semantic Model\footnote{BRAIN2NLP Language Processor.URL: http://demo.brain2.online/\#nlp}.

\section{Method Evaluation}

The approach was tested on the problem of classification of text pairs (paraphrases). The task of classification was to match each pair of text with a class from the set {1 (paraphrase), 0 (not sure), -1 (not paraphrase)}.

\subsection{Dataset}

The following data was used for the model – a gold corpus of 808 paraphrases pairs based on news articles as training data and a corpus of 786 paraphrases pairs as test data. 

The expert's analysis of the data was as follows. We have chosen 400 random paraphrases from classes 0 and -1. Class “1” paraphrases were left out because they were all marked correctly and did not raise any questions with the experts. 
\begin{center}
\begin{table}[h]
\begin{flushleft}
\caption{Examples of questionable classification strings}
\end{flushleft}
\begin{tabular}{|c|l|l|c|c|}
\hline
\begin{tabular}[c]{@{}c@{}}GOLD\\ True\_class\end{tabular} & \multicolumn{1}{c|}{T1} & \multicolumn{1}{c|}{T2} & \begin{tabular}[c]{@{}c@{}}Human \\   (an expert)\end{tabular} & \begin{tabular}[c]{@{}c@{}}GOLD \\ vs \\ Human\end{tabular} \\ \hline
0 & \begin{tabular}[c]{@{}l@{}}Liberal Democrat\\  leader Nick Clegg \\ resigned after failing \\ the election.\end{tabular} & \begin{tabular}[c]{@{}l@{}}The leader of the \\ British labour party \\ resigned because of \\ the defeat in the\\ elections.\end{tabular} & 1 & 0 \\ \hline
0 & \begin{tabular}[c]{@{}l@{}}Vasilyeva's lawyer \\ will appeal against the \\ sentence of five years\\ imprisonment.\end{tabular} & \begin{tabular}[c]{@{}l@{}}Vasilyeva's lawyer is \\ appealing the verdict.\end{tabular} & 1 & 0 \\ \hline
0 & \begin{tabular}[c]{@{}l@{}}Putin dismissed a \\ number of generals.\end{tabular} & \begin{tabular}[c]{@{}l@{}}Vladimir Putin has \\ dismissed from office \\ almost 20 generals.\end{tabular} & 1 & 0 \\ \hline
-1 & \begin{tabular}[c]{@{}l@{}}The results of  the\\ elections in the UK \\ strengthened the pound.\end{tabular} & \begin{tabular}[c]{@{}l@{}}The final results\\ of the elections became\\ known to the public.\end{tabular} & 0 & 0 \\ \hline
-1 & \begin{tabular}[c]{@{}l@{}}Media: Ukrainian\\  boxer angered his fans \\ with St. George ribbon.\end{tabular} & \begin{tabular}[c]{@{}l@{}}Ukrainian boxer\\  removed the photo \\ with St. George \\ ribbon from Instagram.\end{tabular} & 0 & 0 \\ \hline
0 & \begin{tabular}[c]{@{}l@{}}Former nuclear\\   scientist charged in the \\ US for the attempted \\ theft of nuclear \\ weapons.\end{tabular} & \begin{tabular}[c]{@{}l@{}}Former nuclear \\ physicist accused in\\ the US for trying to\\ steal information on \\ nuclear weapons.\end{tabular} & 1 & 0 \\ \hline
0 & \begin{tabular}[c]{@{}l@{}}Vice President of \\ Guatemala resigned \\ due to corruption \\ scandal.\end{tabular} & \begin{tabular}[c]{@{}l@{}}Vice President of\\ Guatemala will leav\\ e his post because of\\  the scandal.\end{tabular} & 1 & 0 \\ \hline
-1 & \begin{tabular}[c]{@{}l@{}}Police prevented three \\ terrorist attacks in \\ Melbourne, Australia,\end{tabular} & \begin{tabular}[c]{@{}l@{}}Terrorists had\\   prepared three\\ terrorist attacks in \\ Melbourne, Australia.\end{tabular} & 1 & 0 \\ \hline
\end{tabular}
\end{table}
\end{center}

\subsection{Numerical experiment}

Calculations were done for:
\begin{itemize}
\item[-] The semantic cores method, where vector representation contains only the semantic part;
\item[-] Alekseev’s method, where vector representation is complemented with semantic and grammar parts;
\item[-] Expert assessment data.
\end{itemize}

For the semantic cores method, we used a measure of texts proximity which was developed earlier for this model \cite{2}. For the Alekseev’s method we used the above-described algorithm for computing the cosine between the two texts (groups of words) as a proximity measure. Thus, both in the first and in the second model, each pair of texts is compared with a real number in the range from 0 to 1 — a measure of proximity.

Furthermore, an algorithm of paraphrase class prediction by the proximity value was used for both models: 

\begin{equation}
Class(\mu )=\begin{cases}1, \ \mu <b \\ 0 \quad a< \mu <b  \\  -1, \ otherwise \end{cases}
\end{equation}
where a,b are clipping constants

We also considered the option of classification into two classes - combining classes 0 and -1 (or classes 1 and 0). In this case, the prediction algorithm is simplified, and there is only one cut-off level remains instead of two.

We have trained a model to find the optimal values of these levels (the target optimization function is precision – the proportion of correctly predicted classes among all predictions). For this purpose, the marked up dataset (1913 records) was supplemented by the calculation of the proximity measure and divided into two parts at a ratio of 70 to 30\%. The bigger part was used as training data, the rest was used for prediction.

The prediction results for the trained models are presented in Table 2.

\begin{table}[h]
\begin{flushleft}
\caption{Methods’ Evaluation}
\end{flushleft}
\begin{tabular}{|l|c|c|c|c|c|c|}
\hline
\multicolumn{1}{|c|}{\multirow{2}{*}{\begin{tabular}[c]{@{}c@{}}Methods and \\ object classes\end{tabular}}} & \multicolumn{2}{c|}{Bound} & \multicolumn{4}{c|}{Evaluation} \\ \cline{2-7} 
\multicolumn{1}{|c|}{} & Lower (a) & Upper (b) & Precision & Recall & F1-score & Accuracy \\ \hline
\begin{tabular}[c]{@{}l@{}}The semantic cores method \\   2 classes: 1 and -1+0\end{tabular} & -- & 0.92 & 0.5705 & 0.5069 & 0.4690 & 0.80 \\ \hline
\begin{tabular}[c]{@{}l@{}}The semantic cores method\\   2 classes: 1+0 and -1\end{tabular} & 0.76 & -- & 0.6825 & 0.5558 & 0.4096 & 0.47 \\ \hline
\begin{tabular}[c]{@{}l@{}}The semantic cores method\\   3 classes: -1, 0, 1\end{tabular} & 0.88 & 0.89 & 0.6182 & 0.3557 & 0.2398 & 0.41 \\ \hline
\begin{tabular}[c]{@{}l@{}}Alekseev’s method\\   2 classes: 1 and -1+0\end{tabular} & -- & 0.96 & 0.7078 & 0.5334 & 0.5169 & 0.81 \\ \hline
\begin{tabular}[c]{@{}l@{}}Alekseev’s method\\   2 classes: 1+0 and -1\end{tabular} & 0.47 & -- & 0.8007 & 0.5043 & 0.3841 & 0.60 \\ \hline
\begin{tabular}[c]{@{}l@{}}Alekseev’s method\\   3 classes: -1, 0, 1\end{tabular} & 0.47 & 0.96 & 0.6736 & 0.3512 & 0.2363 & 0.41 \\ \hline
\begin{tabular}[c]{@{}l@{}}Expert \\   2 classes: 1+0 -1\end{tabular} & -- & -- & 0,8008 & 0,6227 & 0,7032 & 0,61 \\ \hline
\end{tabular}
\end{table}

\begin{center}
\begin{figure}[h!!]
\includegraphics[width=0.6\linewidth]{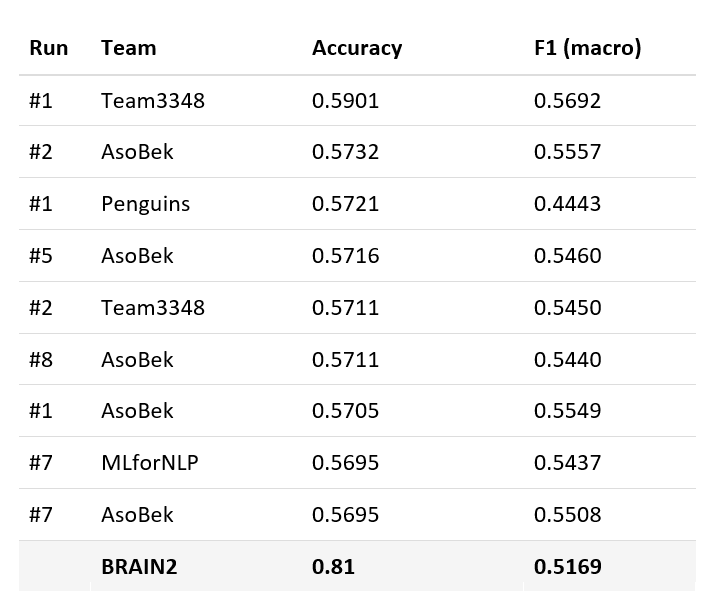}
\caption{Top results for the paraphrase competition track}
\label{tabl}
\end{figure}
\end{center}

We compared the results of our team with publicly available competition results. An online competition track was held based on ParaPhraser corpus. The aim was to train a model to detect sentences which are paraphrases in Russian language. The training corpus available for the participants had 7000 paraphrase paraphrase pairs, test corpus consisted of 1000 pairs. 10 best results appear in the figure \ref{tabl} . 
As we can see the best achieved results scored the accuracy of 0.59 and F1-measure of 0.5692, whereas Alexeev’s method scored 0.81 accuracy and 0.5169 F-measure.

\section*{Conclusion}
The obtained results demonstrate that Alekseev’s method allows solving problems where comparison of vector groups is required. So, in comparison with semantic cores method it provides better classification quality. The presented method of calculating the texts’ proximity measure can be used in practical problems of machine learning.

In the future, we plan to consider the use of this method in other problems that allow vector representation of compared objects. We should also consider the conditions for the singularity of the matrix $a. T \dot a$, and also propose formulas for calculating the measure of proximity in this singularity case. We intend to investigate efficient ways to repeatedly compute the proximity measure of different vectors $b$ with respect to a fixed group of vectors $a$. We plan to exclude the repeated calculation of the middle part (dependent only on the matrix $a$ which represents the group) $\cdot a^T (a \cdot a^T)^{-1} \cdot a$ of the above formula of the angle cosine between vector $b$ and group $a$.

\bibliographystyle{unsrt}

\begin{thebibliography}{1}

\bibitem{1}	A.~Artemov, A.~Sergeev, I.~Khasenevich, A. Yuzhakov, M. Chugunov. The Training of Neuromodels for Machine Comprehension of Text. BRAIN2Text Algorithm. 2018. https://arxiv.org/abs/1804.00551

\bibitem{2} A.Artemov, I.Bolokhov, D.Kem, I.Khasenevich. Neural Network-based Object Classification by Known and Unknown Features (Based on Text Queries). 2019. https://arxiv.org/abs/1906.00800 

\bibitem{3} W. H. Gomaa and A. A. Fahmy, “A survey of text similarity approaches,” Int. J. Comput. Appl., vol. 68, no. 13, 2013, doi: https://doi.org/10.5120/11638-7118. 

\bibitem{4} Mihalcea R. , Corley C. \& Strapparava C. (2006). Corpus-based and Knowledge-based Measures of Text Semantic Similarity. In Proceedings, The Twenty-First National Conference on Artificial Intelligence and the Eighteenth Innovative Applications of Artificial Intelligence Conference, July 16-20, 2006, Boston, Massachusetts, USA.

\bibitem{5} A. Sieg. Text Similarities : Estimate the degree of similarity between two texts. July, 2018. https://medium.com/@adriensieg/text-similarities-da019229c894

\bibitem{6} J. Wang, G. Li, and J. Fe, “Fast-join: An efficient method for fuzzy token matching based string similarity join,”  in  2011  IEEE  27th  International  Conference  on  Data  Engineering,  2011,  pp.  458–469,  doi: https://doi.org/10.1109/ICDE.2011.5767865

\bibitem{7} Wang S., Wang W., Zhuang Y. \& Fei X. (2015). An ontology evolution method based on folksonomy. In Journal of Applied Research and Technology. JART

\bibitem{8} A. Yunianta, O. M. Barukab, N. Yusof, N. Dengen, H. Haviluddin, and M. S. Othman, “Semantic data mapping technology to solve semantic data problem on heterogeneity aspect,” Int. J. Adv. Intell. Informatics, vol. 3, no. 3, pp. 161–172, Dec. 2017, doi: https://doi.org/10.26555/ijain.v3i3.131.
\end{thebibliography}

\end{document}